\documentclass{article}

\usepackage{arxiv}

\usepackage[utf8]{inputenc} 
\usepackage[T1]{fontenc}    
\usepackage{hyperref}       
\usepackage{url}            
\usepackage{booktabs}       
\usepackage{amsfonts}       
\usepackage{nicefrac}       
\usepackage{microtype}      
\usepackage{lipsum}
\usepackage{comment}
\usepackage{times}
\usepackage{latexsym}
\usepackage{mathtools}
\usepackage{float}
\restylefloat{table}
\usepackage{subfig}
\usepackage[dvipsnames]{xcolor} 

\newcommand\R{\mathbb{R}}
\newcommand\N{\mathbb{N}}

\title{About Graph Degeneracy, Representation Learning and Scalability}

\author{
  Simon Brandeis\\
  CentraleSupelec\\
  France \\
  \texttt{simon.brandeis@student.ecp.fr} \\
   \And
 Adrian Jarret \\
  CentraleSupelec\\
  France \\
  \texttt{adrian.jarret@student.ecp.fr} \\
  \AND
  Pierre Sevestre\\
  CentraleSupelec \\
  France \\
  \texttt{pierre.sevestre@student.ecp.fr} \\
}

\begin{document}
\maketitle

\begin{abstract}
Graphs or networks are a very convenient way to represent data with lots of interaction. Recently, Machine Learning on Graph data has gained a lot of traction. In particular, vertex classification and missing edge detection have very interesting applications, ranging from drug discovery to recommender systems. To achieve such tasks, tremendous work has been accomplished to learn embedding of nodes and edges into finite-dimension vector spaces. This task is called \textit{Graph Representation Learning}. However, Graph Representation Learning techniques often display prohibitive time and memory complexities, preventing their use in real-time with business size graphs. In this paper, we address this issue by leveraging a degeneracy property of Graphs - the \textit{K-Core Decomposition}. We present two techniques taking advantage of this decomposition to reduce the time and memory consumption of walk-based Graph Representation Learning algorithms. We evaluate the performances, expressed in terms of quality of embedding and computational resources, of the proposed techniques on several academic datasets.
Our code is available at \href{https://github.com/SBrandeis/k-core-embedding}{https://github.com/SBrandeis/k-core-embedding}.
\end{abstract}


\section{Introduction}
\label{sec:intro}
{
    \subsection{Context and motivations}
    {
    Graph embeddings, \textit{i.e.} graph representations into vector spaces, are a useful representation for many downstream machine learning applications. However, it is crucial to determine what makes a "good" embedding. There exists two main qualities that are often considered as a target for a good embedding when it comes to graphs : structural equivalence and $k^{th}$-order similarity. Structural equivalence states that two vertices of the graph that see the same (or a similar) neighbourhood should have similar embeddings, with respect to the embedding of their common neighbourhood. $k^{th}$-order similarity is the notion of $k^{th}$-neighbour. It simply considers that the embeddings should conserve the proximity in the vector space that existed in the graph. Two nodes related with a short path should be quite close in the embeddings space.
    
    Such things been stated, we also should emphasize that finding such good quality embeddings is a time consuming task. The main focus of our paper is to study ways to speedup the execution time of the embedding generation procedure, while maintaining a good embeddings quality.
    }

    \subsection{Mathematical framework}
    {
    Let us introduce the mathematical notations we will use.
    
        \subsubsection{General notations}
        {
        Let's consider a Graph (\textit{a.k.a.} Network) $G$, composed of a set of vertices or nodes $V$ and a set of edges $E \subset V^2$:
        \begin{equation}
            G = (V, E)
        \label{eq:def_graph}
        \end{equation}
        We will denote by $v$ an element of $V$, that is a vertex or node of $G$. The symbol $e$ will denote an element of $E$, that is an edge of $G$. Each element $e$ of $E$ represents a connection between two nodes, and as such can be represented as a pair $(u, v)$ of elements of $V$. This connection can be unidirectional, meaning the order in the pair $(u, v)$ matters. We will call such connections \textit{directed} edges and will note them: $(u \rightarrow v)$. \textit{A contrario}, an edge can be bidirectional and we will call it \textit{undirected}. A graph will be said \textit{directed} or \textit{undirected} if all of its edges are respectively directed or undirected. Moreover, edges in a Graph can have a relative importance. We will model this by assigning to each edge $(u, v)$ a numerical \textit{weight} $w_{u, v}$. A graph presenting such kind of edges will be called a \textit{weighted} graph. When all edges are equivalent, the graph is said to be \textit{unweighted}.
        
        We can define an inclusion relationship between two graphs $G_1 = (V_1, E_1)$ and $G_2 = (V_2, E_2)$ as follows:
        
        \begin{align}
            &\begin{aligned}
            G_1 \subset G_2 \iff &V_1 \subset V_2\\
                                 &\text{and} \quad E_1 \subset E_2\\
                                 &\text{and} \quad E_1 \subset V_1 \times V_1
            \end{aligned}
            \label{eq:subgraph}
        \end{align}
        
        $G_1$ is then said to be a \textit{subgraph} of $G_2$. The other way around, $G_2$ is a \textit{supergraph}  of $G_1$.
        
        To quantify the connectivity of a node or vertex, we use the notion of \textit{degree} of a node. In the case of undirected and unweighted graphs, the degree of a node is simply the number of edges including this node:
        
        \begin{equation}
            \forall v \in V \quad \quad \text{deg}(v) = |\{(u, v) \quad \forall u \in V : (u, v) \in E \}|
        \end{equation}
        
        Where $|.|$ denotes the cardinality of the set.
        
        The degree of a node can be generalized to directed graph by counting in-bound and out-bound edges separately. The number of in-bound edges and out-bound edges are then respectively called \textit{in-degree} and \textit{out-degree} of the node. When the edges are weighted, the degree is defined to be the sum of the weights corresponding to the aforementioned edges. We can represent the degree of each node in the graph via the diagonal \textit{degree matrix} $D = (d_{u, v})_{(u, v) \in V \times V}$ of size $|V| \times |V|$:
        
        \begin{equation}
            d_{u, v} = \begin{cases}
                         \text{deg}(u) \quad \text{if} \quad  u = v \\
                         0 \quad \quad \quad \text{otherwise}
                      \end{cases}
        \end{equation}
        
        A graph $G$ can be represented by its adjacency matrix $A$, of size $|V| \times |V|$, that defines the connections between the nodes of G. An edge of weight $w_{u, v}$ exists from node $u$ to node $v$ if and only if the element $A_{u, v}$ of coordinates $u, v$ is equal to $w_{u, v}$:
        
       \begin{align}
        &\begin{aligned}
            \forall (u, v) \in V \times V \quad  \quad 
            A_{u, v} &= w_{u, v} \neq 0\iff (u \xrightarrow{w_{u, v}} v) \in E\\
                    &= 0 \iff (u \rightarrow v) \notin E
        \end{aligned}
        \label{eq:adj_mat}
        \end{align}
        }
        
        \subsubsection{Graph Representation Learning}
        {
        \textit{Graph Representation learning} is the task of learning a "good" representation function $f$ that maps each node $v$ of the Graph to a representation in a $n$-dimensional vector space. That is, find:
        
        \begin{equation}
            f: V \rightarrow \R^n
        \label{eq:def_mapping}
        \end{equation} 
        
        that has relevant properties for the task we want to achieve. We can extend this definition to the set of edges $E$ of the graph.\\
        
        The image of $V$ through the mapping $f$ will be called an \textit{embedding} of the graph. We will denote it $X_V$. Its elements will be noted $x_v$, $v$ living in $V$.
        
        \begin{align}
        &\begin{aligned}
        X_V &= im_f(V) \\
            &= \{x_v = f(v) \quad \forall v \in V \} \subset \R^n
        \end{aligned}
        \label{eq:def_embedding}
        \end{align}
        \\
        
        As mentioned above, there exist plenty of criterions to evaluate the quality of an embedding. Although, metrics such as the $k$-th order proximity are hard to verify in the embedded vector space. In order to quantify the quality of the information from the graph retained in the embedding, we trained a logistic regression to solve a downstream classification task - \textit{e.g.} node classification or missing edge detection - on the embedded space. The quality of the embedding is then measured by the $F_1$ score of the classifier:
        
        \begin{equation}
            F_1 = 2\times\frac{\text{precision}\times\text{recall}}{\text{precision} + \text{recall}}
        \label{eq:f1_score}
        \end{equation}
        }
        
        \subsubsection{Degeneracy and $k$-core decomposition}
        {
        Another key concept of our study is the degeneracy of graphs. It is a way to measure the density of graphs. Let $k$ be an integer, a graph $G$ is said to be $k$-degenerate if all its subgraphs have a vertex of degree at most $k$. The smallest $k$ for which the graph is $k$-degenerate is called the \textit{degeneracy} of the graph, and we'll note it $k_{degeneracy}$.
        
        In addition, for every integer $k$ lower than $k_{degeneracy}$, we can define the \textit{$k$-core} of the graph as the \textit{maximal} (in the sense of the inclusion relationship defined in equation \ref{eq:subgraph}) \textit{connected sub-graph} of $G$ where all vertices have a degree of at least $k$. The degeneracy of the graph is then the largest $k$ such that the $k$-core is not empty. 
        
        \begin{align}
            &\begin{aligned}
                &\forall k \leq k_{degeneracy}, \quad \text{core}_G(k) \subset G \\
                &\forall v \in \text{core}_G(k), \quad \text{deg}(v) \geq k \\
                &\forall \text{subgraph} \subset G \quad \text{such that:} \quad \forall v \in \text{subgraph} \quad \text{deg}(v) \geq k , \quad \text{subgraph} \subset  \text{core}_G(k)
            \end{aligned}
            \label{eq:k-core}
        \end{align}
        
        A node of the graph $G$ is said to have a \textit{core number} or \textit{core index} of $c \in \N$ when it is a node of the $c$-core but not of the $(c+1)$-core of $G$.
        
        The $k$-cores of a graph are a way to decompose a graph in a hierarchical sequence of smaller and denser sub graphs. The so-called \textit{$k$-core decomposition} has been used notably for massive graph visualization \cite{alvarez2006large} and Graph structure analysis\cite{nikolentzos2018degeneracy} \cite{giatsidis2014corecluster} purposes.
        }

        \subsubsection{Random walk in a graph}
        {
        Finally, we need to introduce the notion of random walk in a graph. Given a node $v$ of the graph, we can define a stochastic process $W_v$ such that:
        
        \begin{align}
        &\begin{aligned}
        W_v^0 &= v \\
        \forall t \in \N \quad W_v^t &\in \text{neighbours}(W_v^{t-1})
        \end{aligned}
        \label{eq:random_walk_process}
        \end{align}
        }
        
        Where $\text{neighbours}(u)$ is the set of neighbours of a node $u$, that is the set of nodes that have a direct connection to the node $u$:
        
        \begin{equation}
            \forall u \in V \text{,} \quad \text{neighbours}(u) = \{ v \in V : (u \rightarrow v) \in E \}
        \label{eq:neighbourhood}
        \end{equation}
        
        A random walk of length $l$ and rooted in $v$ is a sequence of realizations of $W_v$:
        
        \begin{equation}
            (W_v^0 = v, W_v^1, ..., W_v^{l-1})
        \label{eq:random_walk_realizations}
        \end{equation}
        
        Random walks have been used for community detection in graphs \cite{andersen2006local} or for content recommendation \cite{fouss2007random}.
    }
    
    \subsection{Related Work}
    {
    There exists roughly three families of graph representation learning techniques \cite{goyal2018graph}. Factorization-based methods use the adjacency matrix to embed vectors. Random-walk based methods extract context from a node's neighbourhood with random walks and use methods inspired from Natural Language Processing to compute nodes' embeddings. More recently, new methods leveraging neural networks have been proposed to tackle the task of graph embeddings. Due to the lack of a consensual benchmark dataset and evaluation process, it is hard to objectively compare the quality of the embedding produced by those methods.
    
        \subsubsection{Factorization-based methods}
        {
        Factorization-based methods embed the graph by factorizing a matrix representing the graph. Matrices used are often the adjacency matrix $A$ or the laplacian matrix $L = D - A$. Works worth citing include Locally-Linear Embeddings or LLE\cite{roweis2000nonlinear}, Laplacian Eigenmaps\cite{belkin2002laplacian}, GraRep\cite{cao2015grarep}, HOPE\cite{ou2016asymmetric}. All cited methods have a time complexity in $O(|E| \times n^2)$ - $n$ being the dimension of the embedding space- except for GraRep that has a time complexity in $O(|V|^3)$.
        
        Factorization-based methods require the factorization of often very large matrices. Indeed, real-life graphs often display tens of thousands of nodes.
        }
        
        \subsubsection{Random walks based Frameworks}
        {
        Random walks (see equations \ref{eq:random_walk_process} to \ref{eq:random_walk_realizations}) based embedding methods were introduced by \cite{Perozzi_2014}. They are based on a simple assessment : the distribution law of vertices appearing in short random walks in the graph follows a power law, much like the distribution of words in natural language. This parallelism inspired \cite{Perozzi_2014} to transpose word embedding techniques such as the SkipGram model \cite{mikolov2013efficient} to nodes in a graph. 
        
        The SkipGram model uses neural network to predict a word, given its context (\textit{e.g.}, a sentence). The idea behind \cite{Perozzi_2014} is to extract the context of each node by doing random walks in the graph, thus building "sentences" of nodes. If we run a sufficient number of walks for each node in the graph, the "node corpus" is supposed to hold enough context information for every node in the graph. Then, the SkipGram model is finally applied to build embeddings of nodes using the random walks as their context, the same way it builds embeddings of words with sentences as the context. This framework is known as DeepWalk.
        
        Numerous papers proposed tweaks to enhance the embedding quality of DeepWalk, the most notable certainly bieng Node2Vec \cite{grover2016node2vec}. The main contribution of Node2Vec is the introduction of two hyperparameters allowing to favour breadth-first sampling (BFS) or depth-first sampling (DFS) at each step of the random walk generation, leading to a significant increase in learning quality representations in complex networks.
        
        Random walk based techniques display a time complexity in $O(|V| \times n)$\cite{goyal2018graph}.
        }
        
        \subsubsection{Graph auto-encoders}{
        Deep neural networks are well-known for their ability to learn useful representations of structured data, \textit{e.g.} images. Graphs form no exception to this, and the use of deep neural networks to address network-structured data has gained a lot of traction in the last five years.
        
        Structural deep network embedding (SDNE) \cite{wang2016structural} and Deep neural graph representation (DNGR) \cite{cao2016deep} (time complexities of respectively $O(|V| \times |E|)$ and $O(|V|^2)$) leverage deep auto-encoders to learn a low-dimensional representation of every node's neighbourhood, \textit{i.e.} its corresponding row the adjacency matrix $A$.
        
        Kipf at al.\cite{kipf2016semi} introduced the notion of convolution over a graph, or Graph Convolutional Networks (GCN), later improved by adding an attention mechanism\cite{velivckovic2017graph}. GCNs have been later used as encoders for Graph Auto-Encoders or GAE \cite{kipf2016variational}\cite{schlichtkrull2018modeling} and in the GraphSAGE algorithm \cite{hamilton2017inductive}. Convolution-based embeddings display a time complexity in $O(|E| \times n^2)$
        }
        
        \subsubsection{Propagation based Frameworks}
        {
        The previous methods have time complexities at least linear with respect to the number of nodes or edges of the graph. While they provide satisfactory results in the task of learning embeddings, they can become impracticable when applied to real life networks having millions of vertices and edges. In consequence, people have work on frameworks to accelerate the whole process of extracting embeddings.
        
        Given an input graph to embed, the idea introduced in the HARP framework \cite{chen2017harp} is to first extract a sequence of subgraphs of decreasing sizes. Each subgraph is a coarsened version of the previous one, obtained by merging nodes that share some common characteristics. A small number of iterations leads to really small graphs: 6 to 8 iterations reduce the size of the input graph by 90 to 95 \%. The next step is to use any embedding method to create a representation of the smallest graph, which can be done quite fast and efficiently as it has a reasonable shape. Finally, the embeddings are propagated back to the successive subgraphs. The agnosticism of this framework to the underlying method makes it very interesting and flexible.
        
        The MILE framework \cite{liang2018mile} refined the general process by taking into account weighted edges in the graph coarsening, and by propagating the obtained embeddings using a graph convolutional neural network.
        
        These two frameworks have made the computation of graph embeddings easier for very large graphs, way larger than what was possible with direct graph embedding techniques. As an example, MILE can comfortably scale to graph with 9 million nodes and 40 million edges, where direct methods run out of time and memory on a modern workstation.
        }
    }
}

\section{Proposed Methods}
\label{sec:proposed_method}
{
The main objective of our work was to take advantage of the core decomposition of graphs to speed up graph embedding techniques. We propose and experimentally evaluate two different methods that we designed. We shall notice that all the embedding strategies presented later on apply to connected graphs. As our methods are only based on the structure of the graphs, each connected subgraph of a given graph are independent. For the sake of simplicity, we will always consider the largest connected subgraph. Consequently, we will only consider graph where the $0$-core is equal to the $1$-core. Indeed, nodes in the $0$-core without being in the $1$-core are  nodes that do not have any connection. It is easy to remark that we cannot extract any information from the context of these nodes (they have an empty context) and so proposing a structural-based vector representation for them makes no sense.

    \subsection{Core-Adaptive Random Walks}
    {
    \label{sec:corewalk}
    DeepWalk\cite{Perozzi_2014} and similar embedding techniques use random walks to explore the context of nodes in the graph and extract structural information from it. The intricacy of a vertex's context is strongly related to its degree: indeed, a node with a high degree (a high number of neighbours) have a more complex "context" than a node with only a couple neighbours. This intuition can be qualitatively verified on random walks rooted from the node: the generated random walks will be much more similar to one another when the degree of the node is lower.
    
    We believe that the higher a node's \textit{core index}, the more intricate its context is, and thus more random walks are necessary to extract meaningful structural information from it. Indeed, a node with a high core index is located in a dense portion of the graph: not only it has lots of connections (a high degree), but its neighbours are likely to be very connected as well. Conversely, a node with a small core index should have fewer neighbours, hence reducing the variety of the context.
    
    Thus, instead of running a fixed number of random walks for every vertex of the graph (as it is done with DeepWalk and Node2Vec), we propose to scale this number according to the $k$-core index, running fewer walks for nodes with a low core index. To be more specific, we propose to scale the number of random walks rooted in a given node $v$ in the graph as follows:
    
    \begin{equation}
        n_v = \max (\lfloor n \times \frac{k_v}{k_{degeneracy}} \rfloor , 1)
        \label{eq:core_adaptive_random_walks}
    \end{equation}
    
    Where $n_v$ is the number of random walks rooted in $v$ to generate, $k_v$ is the core index of node $v$, $k_{degeneracy}$ the degeneracy of the graph, $n$ an arbitrary integer representing the maximum number of random walks rooted in $v$ to generate (reached when $v$ is in the $k_{degeneracy}$-core of the graph), and $\lfloor.\rfloor$ represents the integer flooring operator. 
    
    \begin{figure}[h!]
        \centering
        \includegraphics[width=.6\textwidth]{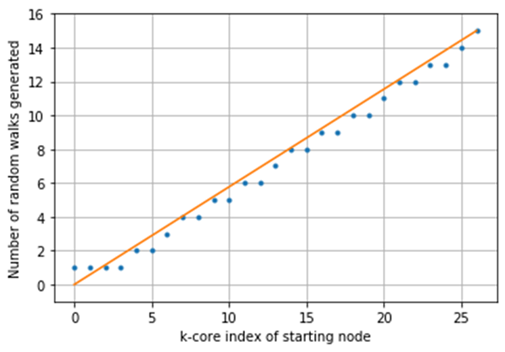}
        \caption{Number of random walks generated (blue dots) versus root core index - $n = 15$ and $k_{degeneracy} = 26$}
        \label{fig:core_adaptive_random_walks}
    \end{figure}
    
    In most of the graphs, the number of nodes per core decreases with the core index, such that the low index cores have a lot of nodes and the high index cores have few to very few nodes. Applying this method reduces drastically the number of random walks to be run for the whole graph, as most of the vertices belong to cores where few random walks are drawn. Even if the task of drawing the random walks can be distributed, and so can be done efficiently, this process of scaling the number of walks still leads to a significant gain in execution time.
    
    We should also notice that the set of all the random walks constitutes the training set of the SkipGram model. With the presented treatment, it get significantly reduced, inducing a way faster training process. However, because the training dataset is smaller, the results could tend to be less accurate. In other words, the quality of the final embedding could be degraded. As stated before, the main goal of this work is to limit this effect as much as possible. We justify our framework with the assumption that running more walks on nodes with simple context would simply add redundant information, and so not running them does not reduce much quality of the embeddings. Additionally, if the quality of the embedding becomes too low, we can use the scaling rule of the walks to increase the size of the dataset. The scaling rule can be used as a parameter to reach a target precision loss in the embedding, compared to a baseline with no reduction of walks.
    }
    
    \subsection{Mean embedding propagation}
    {
    Following the idea of MILE and HARP presented earlier, we propose a framework based on the work presented in \cite{salha2019degeneracy} by \textit{G. Salha et al}. The coarsening procedure we used simply consists in reducing the graph by using the core degeneracy. As the $k$-core subgraphs (also called $k$-degenerates) form a decreasing sequence of graphs (with respect to the inclusion relationship described in \ref{eq:subgraph}), we obtain a usable sequence of "coarsened" graphs. Then we use an embedding algorithm to embed the $k_0$ core subgraph (for a given $k_0$ index, smaller than the maximum $k$ index). For instance, we can use DeepWalk or our algorithm with core adaptive random walks. Then, we use the propagation framework from \cite{salha2019degeneracy} to spread the  embeddings obtained at $k_0$-core to the whole graph.
    
    Their idea is quite simple and shows quality results. Consider that we have built embeddings for the $k$-degenerate graph, and we want to propagate them to the $k-1$-degenerate graph (which is a supergraph of the $k$-degenerate one). Then, each node to embed will be assigned an embedding that is the average of all its neighbors that are either already embedded (from the $k$-degenerate graph) or about to be embedded (from the $k-1$-degenerate graph, but not in the $k-2$-degenerate). We obtain as much equations as there are nodes to be embedded, and as much unknown variables. The new propagated embeddings are the solutions of such a system. To avoid exactly solving the system, \textit{G. Salha et al} introduced an approximation iterative calculus (more information can be found in their paper), which is executed in linear time with respect to the shape of the system (instead of cubic time for exact solution). Eventually, step by step, we propagate up to the whole graph.
    
    To summarize, decomposing the graph into its $k$-core subgraphs is a fast process, even for large graphs. The $k_0$-degenerate subgraph needs to be significantly smaller than the original graph, which is usually true, for $k_0$ large enough. Thus, the embedding step is executed way faster on such subgraph than on the whole graph. Finally, the propagation phase consists in $k_0$ linear steps (with respect to the size of the successive layers of nodes), which is also really fast (compared to the embedding step). The process leads to a significant gain of time, however the propagation step can have consequences on the quality of the final embeddings. To mitigate this impact, one can reduce the value of $k_0$ (the initial degenerate graph to embed with a time expensive strategy). Indeed, with lower $k_0$, the subgraph is larger, and there are less propagation steps, but the embedding step becomes longer. Our contribution has been to use this framework with a random walk based embedding method (DeepWalk).
    }
}

\section{Experiments}
\label{sec:experiments}
{
Let us describe our experimental settings before presenting our results.

    \subsection{Experimental setting}
    {
    \subsubsection{Dataset}
    
    Experiments presented in this paper were performed using three graphs of increasing size: Cora \footnote{https://relational.fit.cvut.cz/dataset/CORA} ($2,708$ nodes, $5,429$ edges) and two graphs from Stanford's SNAP project \footnote{http://snap.stanford.edu/data/index.html}: a subgraph from Facebook ($4,039$ nodes, $88,234$ edges) and a Github developers' connections graph ($37,700$ nodes, $289,003$ edges). These are considered unweighted and undirected.
    
    We plotted the distribution of nodes amongst the different $k$-degenerate subgraphs. Note that $(k+1)$-degenerate subgraph is included within $k$-degenerate, so we plotted the number of nodes that only belong to $k$-degenerate without belonging to $(k+1)$-degenerate, such that each vertex is counted once.

   We should notice that Github graph is quite "regular". The cores nodes distribution follows a behavior that we could reasonably expect : numerous vertices in lowest core indices, decreasing number of nodes with respect to core index, very few in the highest core indices. Cora does not follow the same behavior at all. The graph is quite erratic, with a lot of pairs. Facebook graph is also a bit irregular. It follows the same tendency as Github graph, but the node distribution presents some spikes that are quite unexpected, especially around the 70-core and the last one.

    \subsubsection{Task}
    
    The model is evaluated on a \textit{link prediction} task. For this, one first randomly removes a portion of the edges of the graph ($10 \%$, $30 \%$, and $50 \%$ in our experiments), and trains the model on the resulting graph to obtain the embeddings. Training, validation and test set are obtained using those remove edges as positive samples, and the same number of unconnected pairs of nodes is randomly sampled to account for negative samples. A logistic regression is trained using the concatenation of the embeddings of two nodes as input, predicting the probability of these nodes being connected. 
    
    Performance of the embedding is evaluated using the \textit{F1-score} of the downstream prediction task. Each experiment was repeated 5 times, and the standard deviations are reported along with the results.
    
    Experiments were run using two different CPUs, a \textit{Intel Core i5, 2.4 GHz} with \textit{8 GB} of RAM for the Cora and Facebook networks, and a \textit{Intel Xeon E5-2630 v3, 2.4GHz} with \textit{64 GB} of RAM for the Github graph.
    In the implementation, most graphs operations were performed using \textit{networkx} \footnote{https://networkx.github.io}, and propagation using \textit{sparse scipy} \footnote{https://docs.scipy.org/doc/scipy/reference/sparse.html}.
    
    Additional experiments were performed on the node prediction task, i.e. using the embedding to predict the label of a node for labelled graphs. The model did not provide satisfactory results, suggesting this task requires finer information to perform accurately. 
    
    Main results will be presented in this section, but complete experimental results can be found in the Appendix [\ref{sec:appendix}]. Since the focus of this paper is to study degeneracy-based models to improve random-walk-based embeddings, DeepWalk \cite{Perozzi_2014} will be used as baseline throughout the experiments. 
    
    All DeepWalk parameters are kept to default values, that is 15 random walks per node, of length 30, with a window size of 4. The dimension of the results embedding is 150. 
    
    }
    \subsection{Results}
    
    \subsubsection{Small-size graphs}
    
    For Cora, we apply the propagation framework using DeepWalk as base embedder, on the 2-core up to the $k$-degenerate, that is the $3$-core for the graph after a portion of edges are removed. Experimental results can be found in Table \ref{tab:cora10p}, for \textit{link prediction} with $10\%$ of edges removed. As mentioned before, one would expect the performance to decrease when the initial embedding is performed on a $k_0$-core with increasing $k_0$. Indeed, if the $k_0$ increases, the size of the $k_0$-degenerate graph decreases, making it more difficult for the random-walk based embedder to capture meaningful information, and increasing the propagation task complexity since there are more nodes to propagate the embeddings to. 

        \begin{table}[h!]
    \centering
    \begin{tabular}{@{}llllllll@{}}
    \toprule
    \textbf{Model}    & \multicolumn{2}{l}{\textbf{Performances} (\%)}                & \multicolumn{2}{l}{\textbf{Execution time} (sec.)} \\
    \cmidrule(r{4pt}){2-3} \cmidrule(l){4-5}
             & F1 - Score         & Perf. Drop w.r.t. Baseline & \textbf{Total}           &\textit{Speedup}\\
    \midrule
    \textbf{DeepWalk}   & 58.35 ($\pm$ 1.35) &                          & 37.45 ($\pm$ 2.37)    & \\
    \midrule
    2-core (Dw)          & 58.45 ($\pm$ 0.37) & 0.2                       & 29.77 ($\pm$ 1.62)    & x1.3       \\
    3-core (Dw)          & \textbf{59.21} ($\pm$ 0.9)  & \textbf{1.5}                       & \textbf{14.05} ($\pm$ 0.58)    & \textbf{x2.7 }      \\ \bottomrule
    \end{tabular}
    \caption{\textit{Link prediction} on Cora graph, with $10\%$ of edges removed. Comparison of the DeepWalk baseline with our propagation framework, with DeepWalk as base embedder (K-core(Dw)).}
    \label{tab:cora10p}
    \end{table}

    Surprisingly, on average, this expected performance drop is not observed. However, the standard deviation in each experiment is high, and the overall \textit{F1-score} rather low, making it difficult to draw any conclusion. Yet, a slight speedup of x2.7 is obtained at a reasonable cost. 
    
    \subsubsection{Medium/Large-size graphs}
    
    Similar experiment was performed on the Facebook graph, using DeepWalk to embed the initial $k_0$-core, and performing \textit{link prediction} with increasing values of $k_0$. Results summary can be found in Table \ref{tab:fb10pdw}. As expected, as $k_0$ increases, the $F1$-score decreases, but this drop is at most $11.9\%$ compared to the DeepWalk baseline. On the other side, a significant speedup of x$14.6$ is achieved on the highest core. 
    
   \begin{table}[h!]
    \centering
    \begin{tabular}{@{}llllllll@{}}
    \toprule
    \textbf{Model}    & \multicolumn{2}{l}{\textbf{Performances} (\%)}               & \multicolumn{2}{l}{\textbf{Execution time} (sec.)} \\ 
    \cmidrule(r{4pt}){2-3} \cmidrule(l){4-5} 
             & F1 - Score         & Perf. Drop w.r.t Baseline & \textbf{Total}                  & \textit{Speedup}   \\
    \midrule
    \textbf{DeepWalk} & \textbf{71.67} ($\pm$ 0.33) &                         & 101.92 ($\pm$ 2.93)    &         \\
    \midrule
    9-core (Dw)        & 69.31 ($\pm$ 0.32) & \textbf{-3.3}                      & 70.51 ($\pm$ 0.12)     & x1.4      \\
    25-core (Dw)       & 67.53 ($\pm$ 0.49) & -5.8                      & 48.91 ($\pm$ 3.56)     & x2.1      \\
    49-core (Dw)       & 63.16 ($\pm$ 0.36) & -11.9                     & 21.74 ($\pm$ 1.07)     & x4.7      \\
    73-core (Dw)       & 66.14 ($\pm$ 2.28) & -7.7                      & 7.89 ($\pm$ 0.04)      & x12.9     \\
    97-core (Dw)       & 66.57 ($\pm$ 0.7)  & -7.1                      & \textbf{7} ($\pm$ 0.34)         & \textbf{x14.6}     \\ \bottomrule
    \end{tabular}
    \caption{\textit{Link prediction} on Facebook graph, with $10\%$ of edges removed. Comparison of the DeepWalk baseline with our propagation framework, with DeepWalk as base embedder (K-core(Dw)).}
    \label{tab:fb10pdw}
    \end{table}

    The Core-Adaptive Random walk method (Corewalk) introduced in section \ref{sec:corewalk} was also tested using this graph and the \textit{link prediction} with $10\%$ missing edges. A summary of the results are reported in table \ref{tab:fb10cw}. Alone, without any propagation technique, Corewalk already shows promising results : the $F1$-score obtained is $2\%$ higher than Deepwalk baseline, while having a speedup of x$3$. Consistent results are obtain when removing $30\%$, demonstrating robustness of the solution.
 
    As $k_0$ increases, one observe similar performance drop, at most $11.4\%$ lower than Deepwalk, and the speedup reaches x$13.9$.
    
    \begin{table}[h!]
    \centering
    \begin{tabular}{@{}llllllll@{}}
    \toprule
    \textbf{Model}    & \multicolumn{2}{l}{\textbf{Performances} (\%)}                & \multicolumn{2}{l}{\textbf{Execution time} (sec.)} \\
    \cmidrule(r{4pt}){2-3} \cmidrule(l){4-5}
             & F1 - Score         & Perf. Drop w.r.t. Baseline & \textbf{Total}           &\textit{Speedup}\\
    \midrule
    \textbf{CoreWalk} & \textbf{73.07} ($\pm$ 0.25) &\textbf{ 2}                          & 33.5 ($\pm$ 0.53)      & x3        \\
    9-core (Cw)        & 68.84 ($\pm$ 0.26) & -4                         & 32.97 ($\pm$ 0.86)     & x3.1      \\
    25-core (Cw)       & 68.31 ($\pm$ 0.3)  & -4.7                       & 23.98 ($\pm$ 1.01)     & x4.3      \\
    49-core (Cw)       & 63.51 ($\pm$ 0.48) & -11.4                      & 16.88 ($\pm$ 0.73)     & x6        \\
    73-core (Cw)       & 65.55 ($\pm$ 2.77) & -8.5                       & 8.64 ($\pm$ 0.53)      & x11.8     \\
    97-core (Cw)       & 66.56 ($\pm$ 0.62) & -7.1                       & \textbf{7.34} ($\pm$ 0.95)      & \textbf{x13.9}     \\ \bottomrule
    \end{tabular}
    \caption{\textit{Link Prediction} on Facebook graph, with $10\%$ of edges removed. Comparison of the DeepWalk baseline with the Core-Adaptive Random Walks (CoreWalk) and our propagation framework, with CoreWalk as base embedder. (K-core(Cw))}
    \label{tab:fb10cw}
    \end{table}
    
    These two experiments are summarized in Figure \ref{fig:fb_10} and Figure \ref{fig:fb_30}, for \textit{link prediction} with $10 \%$ and $30\%$ removed edges respectively. First, one can see that the pattern obtained on the $10\%$ task when reaching high core index, \textit{i.e} the increase of \textit{F1-score} with high uncertainty, do not generalize. This pattern is thus most probably due to the particular structure of the graph (\textit{i.e.} important spike around core 60). Then, it is noticeable that the gain in performance using Corewalk do not hold when combined with the propagation technique for high $k_0$. However, one can see that the execution time remains lower until both execution time converge, when the number of nodes to embed initially becomes very low. 
    
    \begin{figure}[h!]
      \centering
      \includegraphics[width=\textwidth]{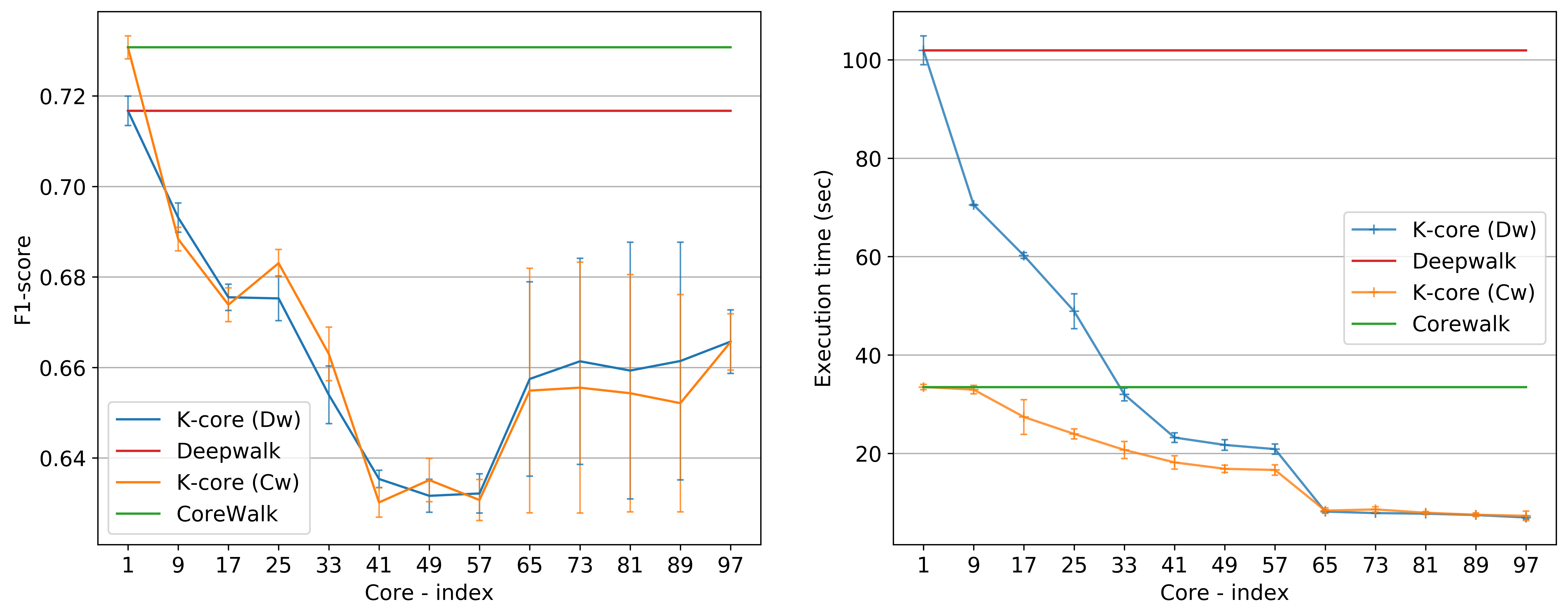}
      \caption{Experiment summary of the \textit{link prediction} task, with $10\%$ of edges removed. (Left) $F1-score$ as a function of the initial embedded core index. (Right) Total execution time as a function of the initial embedded core index.}
    \label{fig:fb_10}   
    \end{figure}
    
    \begin{figure}[h!]
      \centering
      \includegraphics[width=\textwidth]{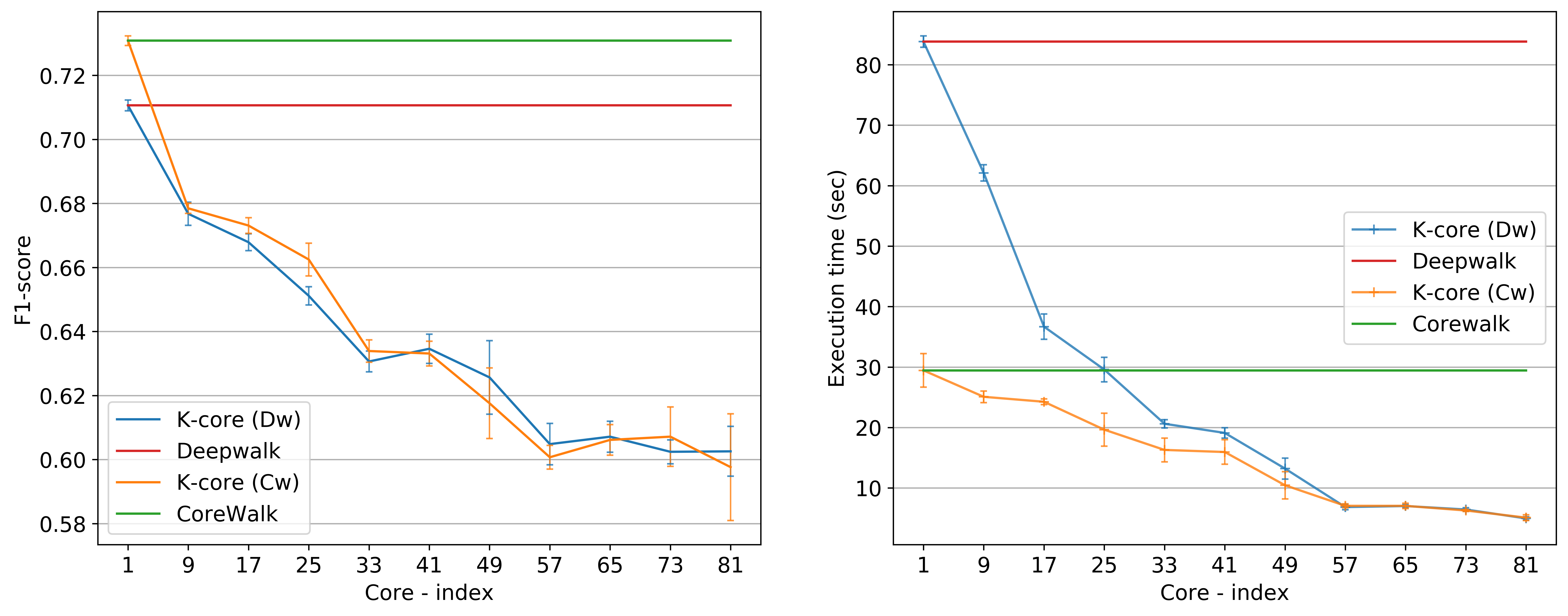}
      \caption{Experiment summary of the \textit{link prediction} task, with $30\%$ of edges removed. (Left) $F1-score$ as a function of the initial embedded core index. (Right) Total execution time as a function of the initial embedded core index.}
    \label{fig:fb_30}
    \end{figure}
    
    Details of this execution time can be found in Figure \ref{fig:exec_time}. The execution time is always dominated by the embedding time of the $k_0$-core, ranging from $100$ to $6$ seconds, when both core decomposition time and propagation time remains below $1$ second. This embedding time is directly proportional to the number of nodes in the corresponding $k_0$-core to embed. 
    
    A final experiment was performed on a Github graph, around 10 times bigger than Facebook, to attest the scalability of the framework. From results displayed in Table \ref{tab:gt10p}, one can notice the sudden drop in performance ($-10.6\%$), even when the initial $k_0$ is rather large (10-core), with a slight speedup (x$3.2$). However, the performance drop lessens for higher core, providing satisfactory results, with around $15\%$ drop in $F1$-score, and an execution time going from more than 10 minutes to around 30 seconds. 
    
    \begin{table}[h!]
    \centering
    \begin{tabular}{@{}llllllll@{}}
    \toprule
    \textbf{Model}    & \multicolumn{2}{l}{\textbf{Performances} (\%)}                & \multicolumn{2}{l}{\textbf{Execution time} (sec.)} \\
    \cmidrule(r{4pt}){2-3} \cmidrule(l){4-5}
             & F1 - Score         & Perf. Drop w.r.t. Baseline & \textbf{Total}           &\textit{Speedup}\\
    \midrule
    \textbf{DeepWalk} & \textbf{74.53} ($\pm$ 0.04) &                          & 684.15 ($\pm$ 4.87)    &      \\
    \midrule
    10-core (Dw)       & 66.71 ($\pm$ 0.2)  & \textbf{-10.5}                      & 214.41 ($\pm$ 1.71)    & x3.2      \\
    20-core (Dw)       & 64.72 ($\pm$ 0.08) & -13.2                      & 84.25 ($\pm$ 2.02)     & x8.1      \\
    30-core (Dw)       & 63.76 ($\pm$ 0.43) & -14.4                      & \textbf{33.19}($\pm$ 0.99)     & \textbf{x20.6 }    \\ \bottomrule
    \end{tabular}
    \caption{\textit{Link prediction} on Github graph, with $10\%$ of edges removed. Comparison of the DeepWalk baseline with our propagation framework, with DeepWalk as base embedder (K-core(Dw)).}
    \label{tab:gt10p}
    \end{table}
    
    \subsubsection{Embedding visualization}
    
    Visualization of the embeddings are provided in the Appendix, in section \ref{sec:vizu}. They are obtained by using Principal Component Analysis (PCA) as dimensional reduction method to project the embeddings onto a 2D space. Two situations are presented : 
    
    \begin{itemize}
        \item The most common case, in Figure \ref{fig:convex_emb}, where the initial embedding is performed on a connected graph. On may notice (\ref{fig:convex_bottom}), that the framework tends to shrink the point cloud to its center in the original projected space of the PCA at each propagation step. The appearing lines in Figure \ref{fig:convex_top} suggests that the framework creates embeddings with very low variance in most dimensions.
        \item Figure \ref{fig:nonconvex_emb} depicts the case when the initial embedding is performed on a $k_0$-core not connected. DeepWalk will then create two distant point clouds, as appearing in \ref{fig:nonconvex_bottom}. The propagation will then put most of the variance in the embedding in this direction to link the two point clouds. This is a drawback of the propagation technique, since the relative positions of the two original point clouds are not related to any property of the whole graph. 
    \end{itemize}
}

\section{Conclusion and Discussion}
\label{sec:conclusion}

Let us recall our guiding principle : accelerate embedding techniques taking advantage of core degeneracy with a limited loss of accuracy. We used downstream tasks to build a metric over the quality of the embeddings produced and DeepWalk as a baseline to compare. A drawback of this approach is that the measure of the embedding quality is biased towards this specific task. To ensure the robustness of the proposed approaches, further work should diversify both the data used (graphs) and the downstream tasks to execute. We still produced promising results that could be an interesting starting step for further work.

Based on random walks algorithms, we proposed a core adaptive framework that scales the number of random walks to explore the nodes' contexts. We observed significant gains of running time on every test graph, with a really moderate drop in prediction tasks. We even observed a slight improvement for larger graphs. Additionally, we reproduced a propagation framework proposed by \textit{G. Salha et al} \cite{salha2019degeneracy}. We used it with DeepWalk and our Core Adaptive Random Walks framework as embedders instead of auto encoders (as they do in their paper). It resulted in an even faster embedding process. Choosing the $k_0$ core index for initial embeddings allows some flexibility to reach a satisfactory loss in performance.

However, the embeddings obtained with such random walks based methods did not lead to outstanding accuracy scores. Indeed, they build graph representations simply using graphs' structural properties. They did not take into account nodes labels or attributes. Thus, our baselines scores are quite low, compared to what can be done with current state of the art techniques (see Graph Convolutional Networks with Attention\cite{velivckovic2017graph} or Graph Attenuated Attention Networks\cite{wang2019knowledge}). DeepWalk was published in 2014, Node2Vec in 2016. To make sure that our frameworks are viable and robust, they should be challenged with more efficient baseline methods as there would be more margin for score loss.

One thing that further work should focus on is connectivity. It is a notion strongly related to the problems we faced during our work, and we believe that improvements can be achieved in this direction. It might happen at some point of the $k$-core decomposition that the obtained subgraph is not connected anymore (imagine a connected graph with two dense areas far one from the other). Then the embedding step creates two clusters of point in the representation space. The mean propagation framework succeeds in connecting them by placing points in between (see figure \ref{fig:nonconvex_bottom}), but the final embeddings are of quite poor quality and do not match well the structure of the graph. An idea to tackle this problem is still to be explored, for instance, we could think of a way to relate not-connected embeddings via generating random walks between the connected areas.

Another way of improvement would be to refine the propagation of the embeddings. In the propagation step, as the index of the $k$-core decreases, a large number of nodes might be added, compared to the number of nodes that have already been assigned an embedding. Thus, we might not have enough information to precisely build nodes representations, and the averaging step could crush together the points. An idea could be to do mean propagation if few nodes are added, and to recompute embeddings if the nodes are too numerous. However, it would be necessary to find a way to compute new embeddings using the ones we already have.

\clearpage
\bibliographystyle{unsrt}
\bibliography{ref}

\clearpage

\section*{Appendix}
\label{sec:appendix}
\subsection{Experimental results}

\begin{table*}[h!]
\resizebox{\textwidth}{!}{\begin{tabular}{@{}llllllll@{}}
\toprule
\textbf{Model} & \multicolumn{2}{l}{\textbf{Performances}(\%)}               & \multicolumn{5}{l}{\textbf{Execution time} (sec.)}                                                          \\ \cmidrule(r{4pt}){2-3} \cmidrule(l){4-8}
      & F1 - Score         & Perf. Drop w.r.t Baseline & Core decomposition  & Propagation  & Embedding  & \textbf{Total}               & \textit{Speedup} \\
      \midrule
\textbf{DeepWalk}          & 58.35 ($\pm$ 1.35) &                         &                &         &      & 37.45 ($\pm$ 2.37) &      \\
\midrule
2-core (Dw)        & 58.45 ($\pm$ 0.37) & 0.2                       & 0.05               & \textbf{0.28}        & 29.44     & 29.77 ($\pm$ 1.62) & x1.3    \\
3-core (Dw)         & \textbf{59.21} ($\pm$ 0.9)  & \textbf{1.5}                       & \textbf{0.03}               & \textbf{0.28}        & \textbf{13.74 }    & \textbf{14.05} ($\pm$ 0.58) & \textbf{x2.7}    \\ \bottomrule
\end{tabular}}
\caption{\textit{Link prediction} on Cora graph, with $10\%$ of edges removed. Comparison of the DeepWalk baseline with our propagation framework, with DeepWalk as base embedder. (K-core(Dw))}
\label{}
\end{table*}

\begin{table*}[h!]
\resizebox{\textwidth}{!}{\begin{tabular}{@{}llllllll@{}}
\toprule
\textbf{Model} & \multicolumn{2}{l}{\textbf{Performances} (\%)}               & \multicolumn{5}{l}{\textbf{Execution time} (sec.)}                                                          \\ \cmidrule(r{4pt}){2-3} \cmidrule(l){4-8}
      & F1 - Score         & Perf. Drop w.r.t Baseline & Core decomposition  & Propagation  & Embedding  & \textbf{Total}               & \textit{Speedup} \\
      \midrule
\textbf{DeepWalk}         & 60.32 ($\pm$ 0.51) &                         &   &       &      & 32.96 ($\pm$ 0.81) &      \\
\midrule
2-core (Dw)          & \textbf{60.41} ($\pm$ 0.94) & \textbf{0.1}                       & 0.07               & \textbf{0.24}        & 23.47     & 23.78 ($\pm$ 1.09) & x1.4    \\
3-core (Dw)          & 59.71 ($\pm$ 0.6)  & -1                        & \textbf{0.02}               & 0.39        & \textbf{5.91 }     & \textbf{6.32} ($\pm$ 0.75)  & \textbf{x5.2}    \\ \bottomrule
\end{tabular}}
\caption{\textit{Link prediction} on Cora graph, with $30\%$ of edges removed. Comparison of the DeepWalk baseline with our propagation framework, with DeeWalk as base embedder. (K-core(Dw))}
\label{}
\end{table*}

\begin{table*}[h!]
\resizebox{\textwidth}{!}{\begin{tabular}{@{}llllllll@{}}
\toprule
\textbf{Model} & \multicolumn{2}{l}{\textbf{Performances} (\%)}               & \multicolumn{5}{l}{\textbf{Execution time} (sec.)}                                                          \\ \cmidrule(r{4pt}){2-3} \cmidrule(l){4-8}
      & F1 - Score         & Perf. Drop w.r.t Baseline & Core decomposition  & Propagation  & Embedding  & \textbf{Total}               & \textit{Speedup} \\
      \midrule
\textbf{DeepWalk}     & 71.67 ($\pm$ 0.33) &                          &                     &            &          & 101.92 ($\pm$ 2.93) &       \\
  \midrule
9-core (Dw)     & 69.31 ($\pm$ 0.32) & -3.3                      & 0.6                     & \textbf{0.56}             & 69.35          & 70.51 ($\pm$ 0.12)  & x1.4    \\
17-core (Dw)    & 67.55 ($\pm$ 0.29) & -5.7                      & 0.58                    & 0.75             & 58.89          & 60.21 ($\pm$ 0.6)   & x1.7    \\
25-core (Dw)    & 67.53 ($\pm$ 0.49) & -5.8                      & 0.48                    & 0.92             & 47.5           & 48.91 ($\pm$ 3.56)  & x2.1    \\
33-core (Dw)    & 65.39 ($\pm$ 0.64) & -8.8                      & 0.41                    & 0.79             & 30.79          & 31.99 ($\pm$ 1.29)  & x3.2    \\
41-core (Dw)    & 63.54 ($\pm$ 0.19) & -11.4                     & 0.33                    & 0.86             & 22.05          & 23.24 ($\pm$ 0.98)  & x4.4    \\
49-core (Dw)    & 63.16 ($\pm$ 0.36) & -11.9                     & 0.32                    & 0.91             & 20.5           & 21.74 ($\pm$ 1.07)  & x4.7    \\
57-core (Dw)    & 63.22 ($\pm$ 0.43) & -11.8                     & 0.31                    & 0.9              & 19.67          & 20.89 ($\pm$ 1.05)  & x4.9    \\
65-core (Dw)    & 65.75 ($\pm$ 2.15) & -8.3                      & 0.19                    & 0.93             & 7.07           & 8.2 ($\pm$ 0.33)    & x12.4   \\
73-core (Dw)    & 66.14 ($\pm$ 2.28) & -7.7                      & 0.19                    & 0.92            & 6.78           & 7.89 ($\pm$ 0.04)   & x12.9   \\
81-core (Dw)    & 65.93 ($\pm$ 2.84) & -8                        & 0.19                    & 0.94             & 6.64           & 7.77 ($\pm$ 0.15)   & x13.1   \\
89-core (Dw)    & 66.14 ($\pm$ 2.63) & -7.7                      & \textbf{0.18}                    & 0.96             & 6.35           & 7.49 ($\pm$ 0.17)   & x13.6   \\
97-core (Dw)    & 66.57 ($\pm$ 0.7)  & -7.1                      & \textbf{0.18}                    & 0.93             & \textbf{5.89}           & \textbf{7} ($\pm$ 0.34)      & \textbf{x14.6}  \\
  \midrule
    \midrule
\textbf{CoreWalk}     & \textbf{73.07} ($\pm$ 0.25) & \textbf{2}                         &        0.65            &            &    32.85       & 33.5 ($\pm$ 0.53)   & x3      \\
  \midrule
9-core (Cw)     & 68.84 ($\pm$ 0.26) & -4                        & 0.62                    & 0.62             & 31.73          & 32.97 ($\pm$ 0.86)  & x3.1    \\
17-core (Cw)    & 67.39 ($\pm$ 0.37) & -6                        & 0.61                    & 0.76             & 26.05          & 27.41 ($\pm$ 3.53)  & x3.7    \\
25-core (Cw)    & 68.31 ($\pm$ 0.3)  & -4.7                      & 0.45                    & 0.82             & 22.7           & 23.98 ($\pm$ 1.01)  & x4.3    \\
33-core (Cw)    & 66.3 ($\pm$ 0.59)  & -7.5                      & 0.37                    & 0.9              & 19.45          & 20.73 ($\pm$ 1.73)  & x4.9    \\
41-core (Cw)    & 63.02 ($\pm$ 0.33) & -12.1                     & 0.33                    & 1.02             & 16.84          & 18.19 ($\pm$ 1.35)  & x5.6    \\
49-core (Cw)    & 63.51 ($\pm$ 0.48) & -11.4                     & 0.32                    & 1.08             & 15.48          & 16.88 ($\pm$ 0.73)  & x6      \\
57-core (Cw)    & 63.07 ($\pm$ 0.45) & -12                       & 0.31                    & 1.1              & 15.24          & 16.65 ($\pm$ 1.06)  & x6.1    \\
65-core (Cw)    & 65.49 ($\pm$ 2.7)  & -8.6                      & 0.2                     & 1.03             & 7.16           & 8.39 ($\pm$ 0.54)   & x12.2   \\
73-core (Cw)    & 65.55 ($\pm$ 2.77) & -8.5                      & 0.2                     & 1.21             & 7.22           & 8.64 ($\pm$ 0.53)   & x11.8   \\
81-core (Cw)    & 65.43 ($\pm$ 2.62) & -8.7                      & 0.19                    & 1.05             & 6.73           & 7.96 ($\pm$ 0.26)   & x12.8   \\
89-core (Cw)    & 65.21 ($\pm$ 2.4)  & -9                        & \textbf{0.18}                    & 1.01             & 6.35           & 7.55 ($\pm$ 0.37)   & x13.5   \\
97-core (Cw)    & 66.56 ($\pm$ 0.62) & -7.1                      & \textbf{0.18 }                   & 1.04             & 6.11           & 7.34 ($\pm$ 0.95)   & x13.9   \\ \bottomrule
\end{tabular}}
\caption{\textit{Link prediction} on Facebook graph, with $10\%$ of edges removed. Comparison of the DeepWalk baseline with the Core-Adaptive Random Walks (CoreWalk) and our propagation framework, with DeepWalk and CoreWalk as base embedder. (K-core(Dw) and K-core(Cw) respectively)}
\label{}
\end{table*}

\begin{table*}[h!]
\resizebox{\textwidth}{!}{\begin{tabular}{@{}llllllll@{}}
\toprule
\textbf{Model} & \multicolumn{2}{l}{\textbf{Performances} (\%)}               & \multicolumn{5}{l}{\textbf{Execution time} (sec.)}                                                          \\ \cmidrule(r{4pt}){2-3} \cmidrule(l){4-8}
      & F1 - Score         & Perf. Drop w.r.t Baseline & Core decomposition time & Propagation time & Embedding time & \textbf{Total}               & \textit{Speedup} \\
      \midrule
\textbf{DeepWalk}     & 71.06 ($\pm$ 0.17) &                         &                     &           &          & 83.83 ($\pm$ 0.95) &      \\
\midrule
9-core (Dw)     & 67.68 ($\pm$ 0.36) & -4.8                      & 0.44                    & \textbf{0.6}              & 61.06          & 62.1 ($\pm$ 1.35)  & x1.3    \\
17-core (Dw)    & 66.79 ($\pm$ 0.27) & -6                        & 0.37                    & 0.74             & 35.57          & 36.68 ($\pm$ 2.09) & x2.3    \\
25-core (Dw)    & 65.12 ($\pm$ 0.29) & -8.4                      & 0.32                    & 0.76             & 28.51          & 29.59 ($\pm$ 2.03) & x2.8    \\
33-core (Dw)    & 63.06 ($\pm$ 0.32) & -11.3                     & 0.27                    & 0.79             & 19.57          & 20.62 ($\pm$ 0.67) & x4.1    \\
41-core (Dw)    & 63.46 ($\pm$ 0.46) & -10.7                     & 0.25                    & 0.82             & 18.04          & 19.11 ($\pm$ 0.86) & x4.4    \\
49-core (Dw)    & 62.57 ($\pm$ 1.15) & -12                       & 0.2                     & 0.81             & 12.2           & 13.21 ($\pm$ 1.73) & x6.3    \\
57-core (Dw)    & 60.48 ($\pm$ 0.65) & -14.9                     & 0.14                    & 0.93             & 5.77           & 6.84 ($\pm$ 0.43)  & x12.3   \\
65-core (Dw)    & 60.71 ($\pm$ 0.48) & -14.6                     & 0.14                    & 0.89             & 5.98           & 7.02 ($\pm$ 0.27)  & x11.9   \\
73-core (Dw)    & 60.24 ($\pm$ 0.37) & -15.2                     & 0.14                    & 0.88             & 5.42           & 6.44 ($\pm$ 0.28)  & x13     \\
81-core (Dw)    & 60.26 ($\pm$ 0.78) & -15.2                     & \textbf{0.13}                    & 0.83             & \textbf{4.01}           & \textbf{4.96} ($\pm$ 0.27)  &\textbf{ x16.9 }  \\
\midrule
\midrule
\textbf{CoreWalk}     & \textbf{73.08} ($\pm$ 0.15) & \textbf{2.8}                      & 0.5                     &              & 28.73          & 29.45 ($\pm$ 2.76) & x2.8    \\
\midrule
9-core (Cw)     & 67.85 ($\pm$ 0.16) & -4.5                      & 0.45                    & 0.64             & 23.99          & 25.08 ($\pm$ 0.95) & x3.3    \\
17-core (Cw)    & 67.31 ($\pm$ 0.24) & -5.3                      & 0.39                    & 0.76             & 23.12          & 24.26 ($\pm$ 0.46) & x3.5    \\
25-core (Cw)    & 66.25 ($\pm$ 0.51) & -6.8                      & 0.32                    & 0.88             & 18.44          & 19.64 ($\pm$ 2.71) & x4.3    \\
33-core (Cw)    & 63.39 ($\pm$ 0.35) & -10.8                     & 0.26                    & 0.97             & 15.06          & 16.29 ($\pm$ 1.97) & x5.1    \\
41-core (Cw)    & 63.31 ($\pm$ 0.39) & -10.9                     & 0.25                    & 1                & 14.69          & 15.95 ($\pm$ 2.03) & x5.3    \\
49-core (Cw)    & 61.76 ($\pm$ 1.1)  & -13.1                     & 0.19                    & 0.91             & 9.35           & 10.45 ($\pm$ 2.25) & x8      \\
57-core (Cw)    & 60.07 ($\pm$ 0.37) & -15.5                     & 0.15                    & 0.87             & 6.03           & 7.05 ($\pm$ 0.31)  & x11.9   \\
65-core (Cw)    & 60.61 ($\pm$ 0.48) & -14.7                     & 0.15                    & 0.97             & 5.93           & 7.04 ($\pm$ 0.43)  & x11.9   \\
73-core (Cw)    & 60.72 ($\pm$ 0.93) & -14.6                     & 0.15                    & 0.87             & 5.26           & 6.28 ($\pm$ 0.16)  & x13.4   \\
81-core (Cw)    & 59.76 ($\pm$ 1.67) & -15.9                     &\textbf{ 0.13}                    & 0.84             & 4.13           & 5.09 ($\pm$ 0.46)  & x16.5   \\ \bottomrule
\end{tabular}}
\caption{\textit{Link prediction} on Facebook graph, with $30\%$ of edges removed. Comparison of the DeepWalk baseline with the Core-Adaptive Random Walks (CoreWalk) and our propagation framework, with DeepWalk and CoreWalk as base embedder. (K-core(Dw) and K-core(Cw) respectively)}
\label{}
\end{table*}

\begin{table*}[h!]
\resizebox{\textwidth}{!}{\begin{tabular}{@{}llllllll@{}}
\toprule
\textbf{Model} & \multicolumn{2}{l}{\textbf{Performances} (\%)}               & \multicolumn{5}{l}{\textbf{Execution time} (sec.)}                                                          \\ \cmidrule(r{4pt}){2-3} \cmidrule(l){4-8}
      & F1 - Score         & Perf. Drop w.r.t Baseline & Core decomposition  & Propagation  & Embedding  & \textbf{Total}               & \textit{Speedup} \\
      \midrule
\textbf{DeepWalk}          & \textbf{74.53} ($\pm$ 0.04) &                           &              &        &    & 684.15 ($\pm$ 4.87) &       \\
\midrule
10-core (Dw)         & 66.71 ($\pm$ 0.2)  &\textbf{ -10.5}                      & 2.85               & \textbf{18.45}       & 193.11    & 214.41 ($\pm$ 1.71) & x3.2    \\
20-core (Dw)         & 64.72 ($\pm$ 0.08) & -13.2                      & 2.23               & 21.07       & 60.95     & 84.25 ($\pm$ 2.02)  & x8.1    \\
30-core (Dw)        & 63.76 ($\pm$ 0.43) & -14.4                      & \textbf{1.84}               & 20.58       & \textbf{10.77}     & \textbf{33.19} ($\pm$ 0.99)  & \textbf{x20.6}   \\ \bottomrule
\end{tabular}}
\caption{\textit{Link prediction} on Github graph, with $10\%$ of edges removed. Comparison of the DeepWalk baseline with our propagation framework, with DeepWalk as base embedder. (K-core(Dw))}
\label{}
\end{table*}

\begin{table*}[h!]
\resizebox{\textwidth}{!}{\begin{tabular}{@{}llllllll@{}}
\toprule
\textbf{Model} & \multicolumn{2}{l}{\textbf{Performances} (\%)}               & \multicolumn{5}{l}{\textbf{Execution time} (sec.)}                                                          \\ \cmidrule(r{4pt}){2-3} \cmidrule(l){4-8}
      & F1 - Score         & Perf. Drop w.r.t Baseline & Core decomposition  & Propagation  & Embedding  & \textbf{Total}               & \textit{Speedup} \\
      \midrule
\textbf{DeepWalk}          & \textbf{71.92} ($\pm$ 0.17) &                           &              &      &   & 631.97 ($\pm$ 2.96) &      \\
\midrule
10-core (Dw)         & 65.27 ($\pm$ 0.26) & \textbf{-9.2 }                      & 1.92               & \textbf{17.55}       & 137.77    & 157.24 ($\pm$ 3.4)  & x4      \\
20-core (Dw)         & 64.45 ($\pm$ 0.28) & -10.4                      & \textbf{1.42  }             & 18.9        & \textbf{27.65}     & \textbf{47.97} ($\pm$ 0.39)  & \textbf{x13.2 } 
\end{tabular}}
\caption{\textit{Link prediction} on Github graph, with $30\%$ of edges removed. Comparison of the DeepWalk baseline with our propagation framework, with DeepWalk as base embedder. (K-core(Dw))}
\label{}
\end{table*}

\subsection{Detailed execution time}
{
    \begin{figure}[H]
      \centering
      \includegraphics[width=\textwidth]{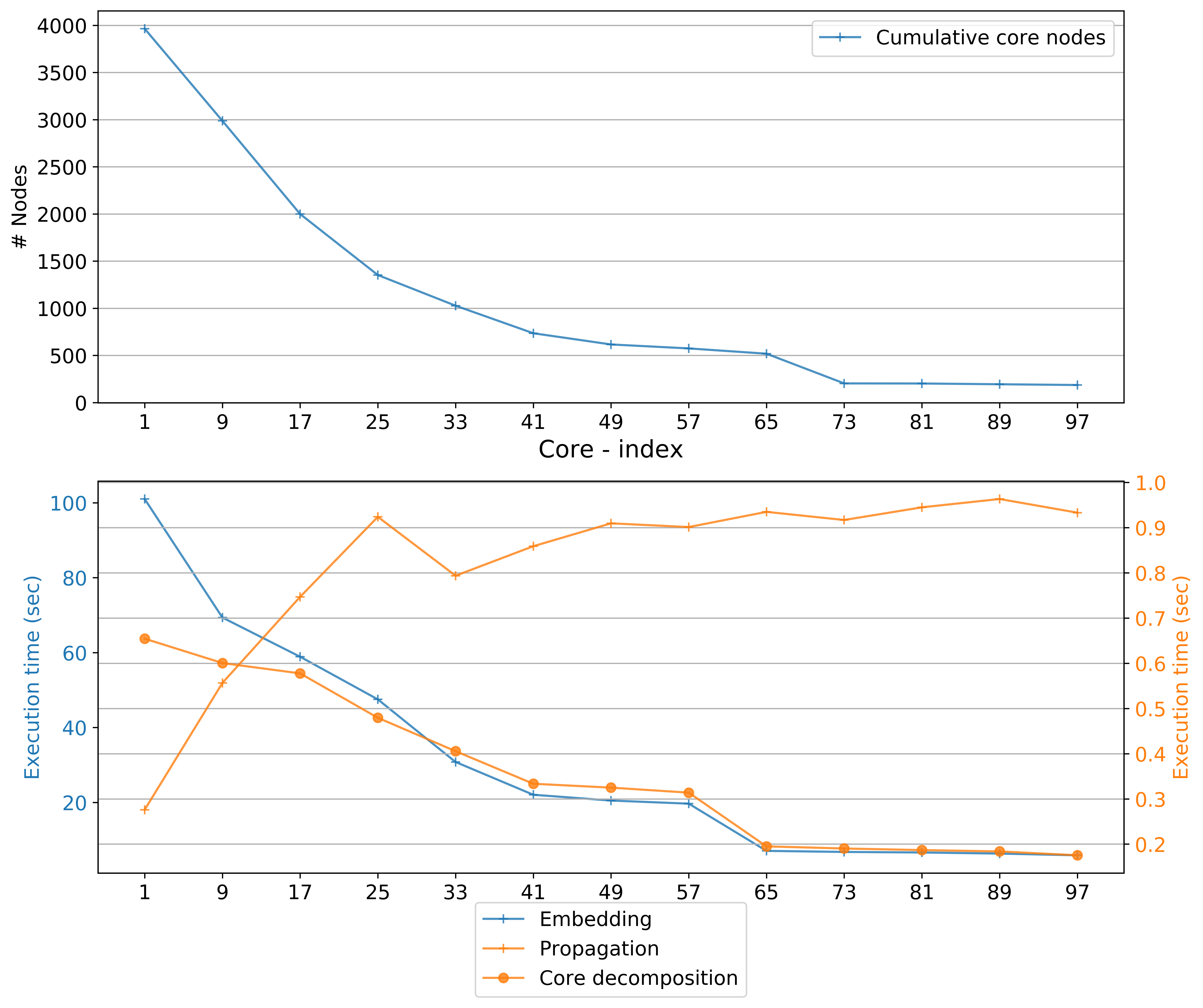}
      \caption{Execution time summary for the \textit{link prediction} task on Facebook graph, $10\%$ of edges removed. (Top) Number of nodes in the initial k-core to embed. (Bottom) Execution time details as a function of the initial $k$-core index.}
    \label{fig:exec_time}
    \end{figure}
}

\subsection{Embedding visualization}
{
\label{sec:vizu}

\begin{figure}[H]%
    \centering
    \subfloat[PCA fitted on the final embedding of the whole graph]{{\includegraphics[width=1\textwidth]{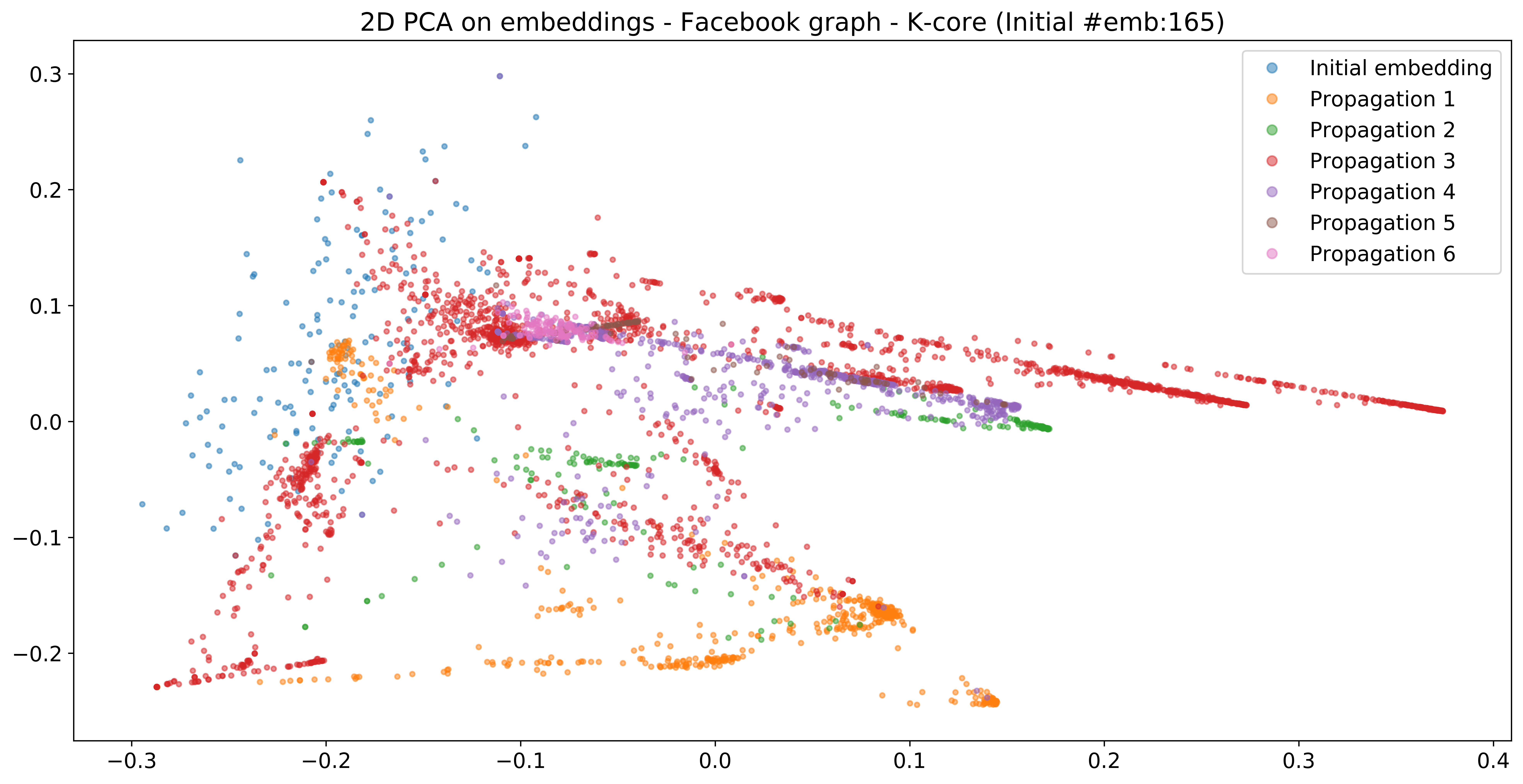}}\label{fig:convex_top}}%
    \qquad
    \subfloat[PCA fitted to the embeddings of the initial $103$-core]{{\includegraphics[width=1\textwidth]{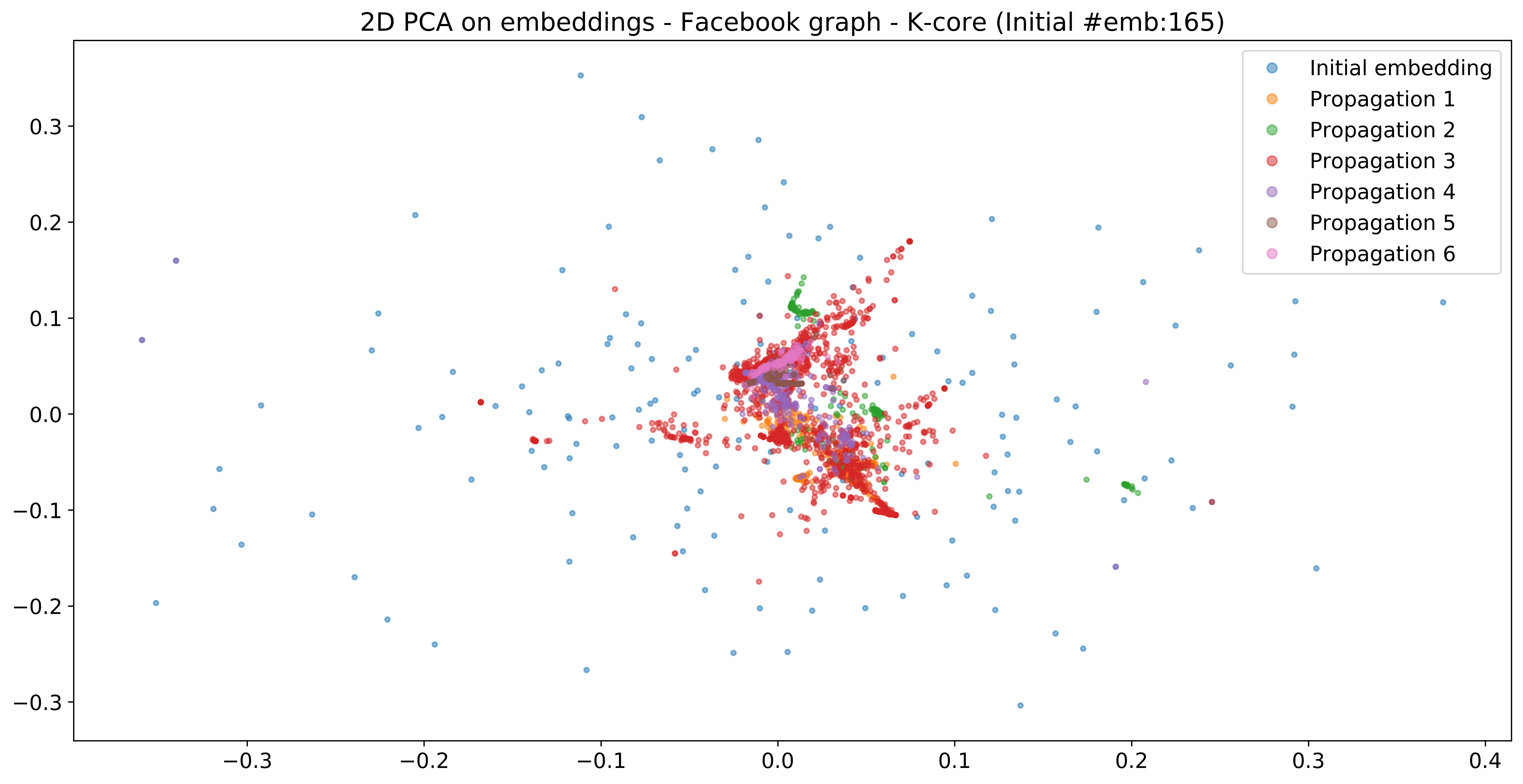}\label{fig:convex_bottom}}}%
    \caption{Visualization of the embeddings obtained on Facebook graph, with $10\%$ of edges removed, and initial embedding performed on the connected $103$-core}%
    \label{fig:convex_emb}%
\end{figure}
    
\begin{figure}[H]%
    \centering
    \subfloat[PCA fitted on the final embedding of the whole graph]{{\includegraphics[width=1\textwidth]{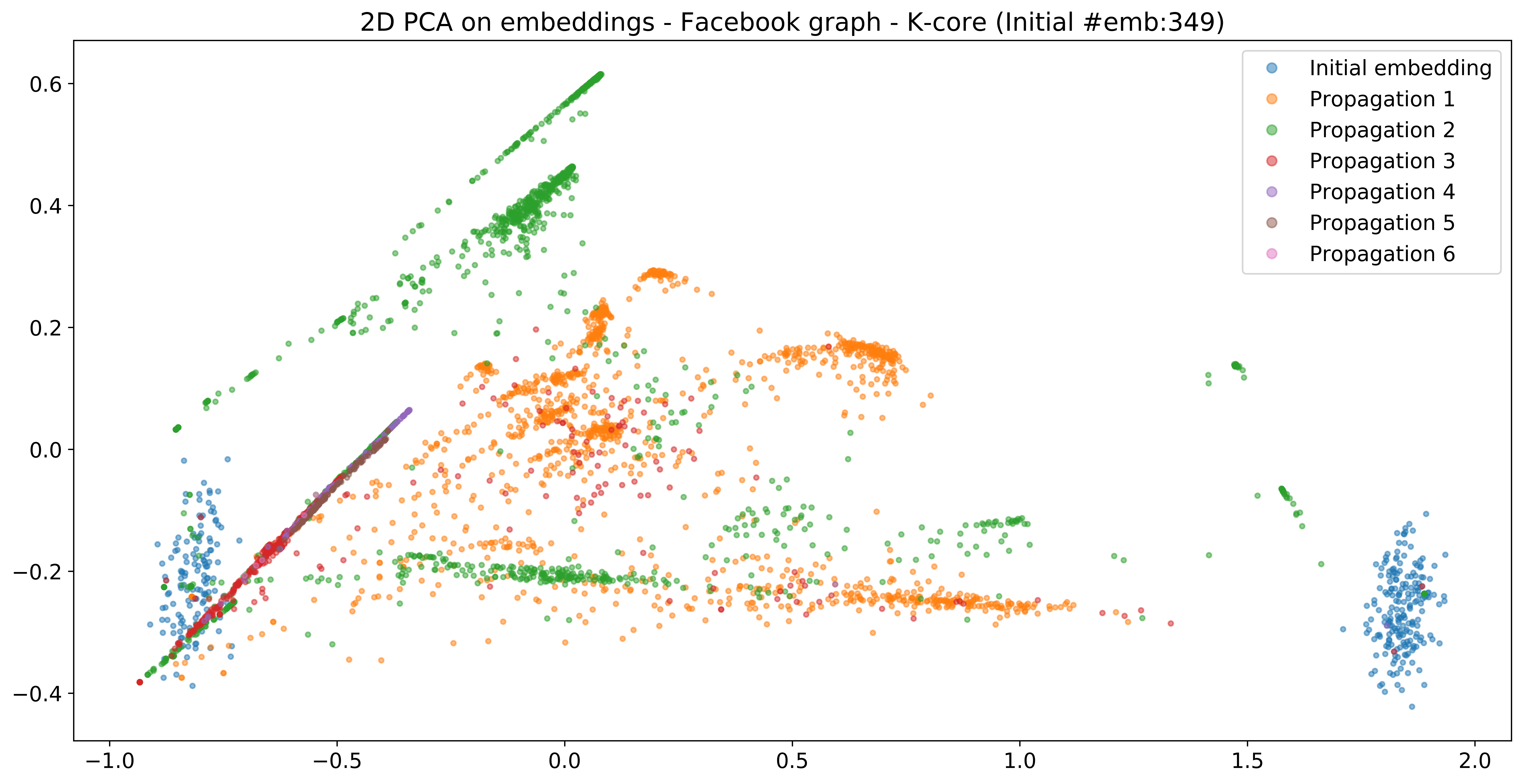}}\label{fig:nonconvex_top}}%
    \qquad
    \subfloat[PCA fitted to the embeddings of the initial $64$-core]{{\includegraphics[width=1\textwidth]{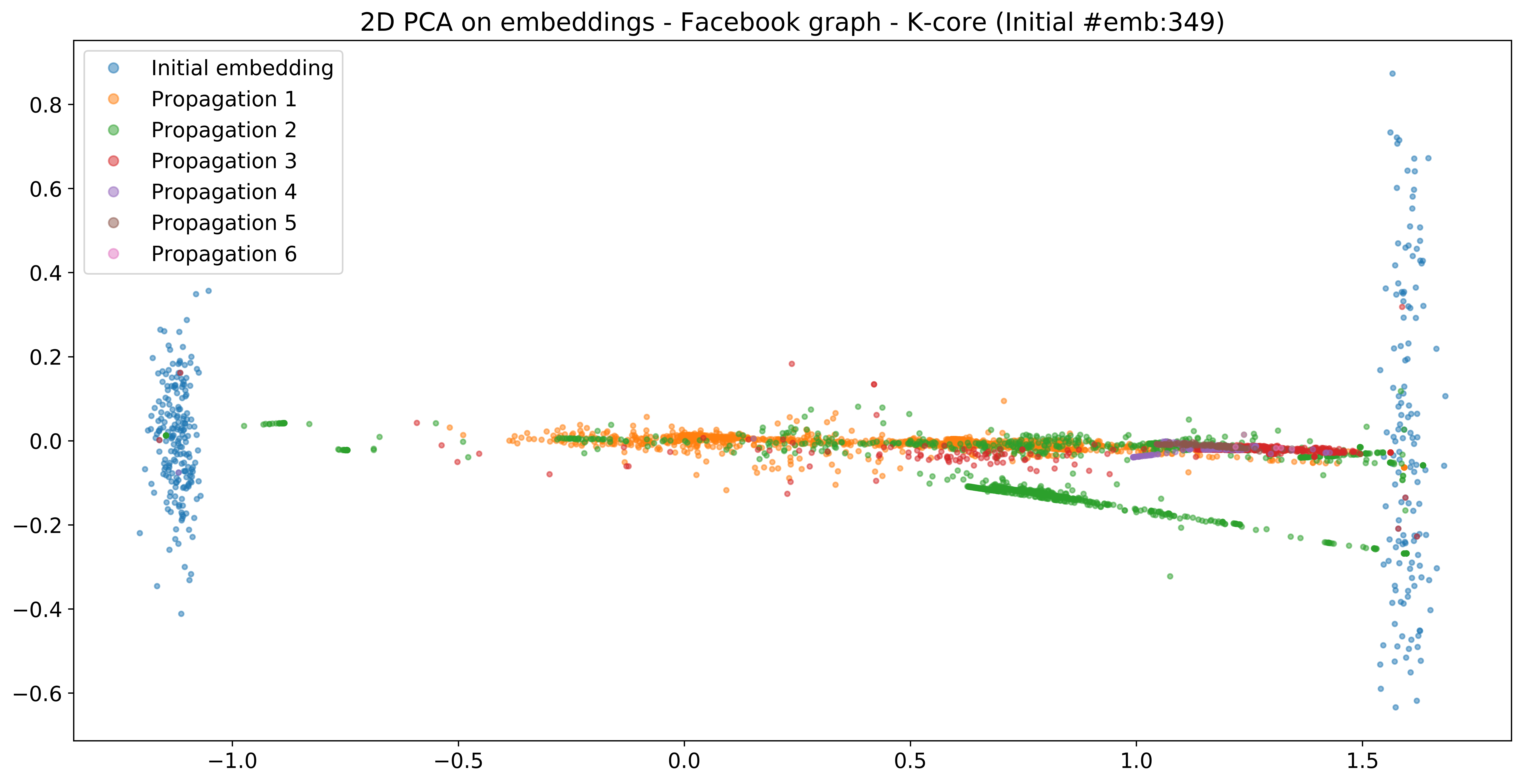}\label{fig:nonconvex_bottom}}}%
    \caption{Visualization of the embeddings obtained on Facebook graph, with $10\%$ of edges removed, and initial embedding performed on the not-connected $64$-core}%
    \label{fig:nonconvex_emb}%
\end{figure}
}

\end{document}